# Computing Machinery and Knowledge

**Raymond Anneborg**

raymond.anneborg@existentialrisk.ai

**Abstract:** The purpose of this paper is to discuss the possibilities for computing machinery, or AI agents, to know and to possess knowledge. This is done mainly from a virtue epistemology perspective and definition of knowledge. However, this inquiry also shed light on the human condition, what it means for a human to know, and to possess knowledge. The paper argues that it is possible for an AI agent to know and examines this from both current state-of-the-art in artificial intelligence as well as from the perspective of what the future AI development might bring in terms of superintelligent AI agents.

## Introduction

In this paper, I will examine virtue epistemology from the perspective of a non-human artificial intelligence (AI) agent to see whether such agent, computing machine, can be able to know things and to possess knowledge. The aim is to gain insight into what it means for a human agent to know things and to possess knowledge by comparing it to a non-human agent in this way.

Alan Turing, one of the founding fathers of AI, wrote in his classical paper "Computing Machinery and Intelligence" (Turing, 1950) about machine intelligence and ask whether computing machinery, digital computers, could be said to think or not. The paper also covers the ability for a machine to learn things. Turing's position was that this was possible, maybe not at his time, but that it would be possible by the end of the century. Well, the end of the century has now passed and today there is a big hype around AI and specifically related to machine learning. My intention is to review selected parts from Alan Turing's paper and put it to the test against virtue epistemology to see if our progress in the field of AI has changed anything in relation to the possibility for machines to think and know things.

My hypothesis, like that of Alan Turing, is that this is possible. The question is if we are there yet or if we need to wait for another end of the century!

## What is knowledge?

What is knowledge? This is the question that the whole field of Epistemology is trying to answer. It might be hard to provide an answer to that question in this short paragraph! I will however propose a definition of knowledge to be used for the purpose of this paper.

Now, as a definition of knowledge I will use the definition from virtue epistemology as proposed by Ernest Sosa:

**"Knowledge is a form of action, to know is to act, and knowledge is hence subject to a normativity distinctive of action, including intentional action."** (Sosa, 2017, pp. 207)

One can argue whether this is a correct definition of knowledge or not, and I am not personally sure that I agree with the definition as stated, however, for the purpose of this paper let us assume that it is correct.

To give us a bit more to work with in this paper when we inquiry into whether a computing machine can know let us elaborate the above definition of knowledge a bit further. Sosa proposes to break down knowledge as action into an AAA and SSS structure where knowledge as action must be Accurate, Adroit and Apt (Sosa, 2017, pp. 72). Accurate meaning it is a successful attempt, adroit meaning an exercise of competence and apt meaning that an attempt was accurate because it was adroit. In an epistemic context, the action related to knowledge become affirmation aimed at truth (Sosa, 2017, pp. 100). For the adroit part related to competence, Sosa then breaks competence down into his SSS structure where competence is constituted by Seat/Skill, Shape and Situation (Sosa, 2017, pp. 191-192). Seat or skill is the innermost in our brain and body, which we always retain. Shape refers to our personal condition i.e. that we are awake and sober. Situation is the external conditions i.e. that it is dark outside and raining. To conclude this elaboration Sosa also talks about first (animal) order and second (reflective) order of knowledge. On the first order, one can affirm aptly, but to know full well one also has to reflect on and judge one's affirmation, and not just make a guess and hope to be lucky, and do so aptly (second order).

## What is computing machinery?

Alan Turing in his paper decided not to define the terms machine and think. Instead, he proposed what he called the imitation game, also known as the Turing Test. Here an interrogator is supposed to ask questions to two agents, one human and one being a digital computer (Turing's term). If the interrogator cannot decide which agent is which, then the proposition is that machines can think. It can be interesting to note that it is still a challenge today to be able to device such computing machine that can pass this test!

Now, for the purpose of this paper I will not propose any test, but instead use a definition of computing machinery that is more up to date and in the context of Artificial Intelligence. Therefore, for this paper I will use the definition that a computing machine is an AI agent. With an AI agent, I will refer to a system built to act intelligent and rational with the objective to achieve the best-expected outcome (see also Russell and Norvig, 2020, chapter 2 – Intelligent Agents, pp. 36-62).

## Can machines know?

So, now with computing machinery and knowledge defined let us address the main question for this paper: "Can machines know?"

To detail the question a bit let us see what this would imply given the virtue epistemology account of knowledge above. For an AI agent to know it would have to be able to perform epistemic actions, alethic affirmations, fully apt. This implies an AI agent has to be accurate in its attempt to affirm, its accuracy has to be adroit meaning competent based on skill, shape and situation, and thus apt. However, this is not enough, the AI agent also has to be able to reflect on its apt affirmation and pass judgement aptly. If we can prove that an AI agent can achieve all of the above, then we should be able to conclude that a machine can know!

Before we go deeper into this inquiry, I would like to bring up one quote from what Turing wrote about "Learning Machines" (Turing, 1950, pp. 17-22):

"We may hope that machines will eventually compete with men in all purely intellectual fields. But which are the best ones to start with? Even this is a difficult decision. Many people think that a very abstract activity, like the playing of chess, would be best. It can also be maintained that it is best to provide the machine with the best sense organs that money can buy, and then teach it to understand and speak English." (Turing, 1950, pp. 17)

This has some interesting points that are worth to bring up to our contemporary time! Today there is a lot of hype around AI and specifically related to machine learning. IBM's AI system Deep Blue in 1996 won against the world champion Garry Kasparov in chess and repeated the feat in 1997! In AI today we have computer vision systems and voice recognition systems that can "see" and "hear", i.e. have cognitive faculties (sense organs/sensors). Finally, we also have natural language systems that can indeed understand and speak English! Therefore, we have today come quite far in the field of artificial intelligence. Can we though, based on this progress, say that machines can know?

Let us start our inquiry with an example. Let us say that we have a person that want to drive safely from work to home. In order to do so this person would first need the <u>skill</u> of driving that is to say that the person would know-how to drive. Secondly, the person would need to be in good <u>shape</u>, not too tired or drunk. The <u>situation</u> would also have to be in suitable condition i.e. if it would be pitch dark outside, there were no streetlights and the headlights of the car did not work, then it would be hard to drive home safely. Therefore, given this SSS-scenario the person would have to reflect on his or her competence to be able to get home safe and pass judgement of whether to drive or walk home. This second order reflection and judgement would then lead to a, hopefully, <u>accurate</u> attempt to get home safe based on the persons competence meaning the successful attempt would be <u>adroit</u> and thus <u>apt</u>, and since it was based on an apt judgement it would be fully apt and thus a show of knowledge by this person.

Now, let us say we have an autonomous vehicle instead, which is an example of an AI agent, which we setup to attempt the same thing to get from work to home in a safe manner. Autonomous vehicles are already today driving in our streets, with examples from Tesla as well as Uber, providing transport services where a passenger can set the vehicle in autopilot, then sit back, and relax while the car does the driving. This is something that will soon be the norm rather than the exception. Do such autonomous vehicles possess knowledge? Let us see! It would need the <u>skill</u> to drive. This is something that it has gained through machine learning algorithms with the objective to maximize safe driving. These machine learning algorithms have been exposed to massive amounts of traffic situations both in simulators and through actual driving in the streets (maybe guided by a human driver) where it has been trained with big volumes of data from sensors (cognitive faculties for machines) like cameras and radar. If you observe cars, you can almost detect if a specific vehicle is on autopilot since it would drive dead center in its lane, while a human driver would wobble back and forth like a bumper car! Therefore, we can say that an autonomous vehicle does indeed have the skill to drive, often better than a human driver does. Let us then test for <u>shape</u>. The autonomous vehicle would of course need to be in good condition, no flat tires or empty fuel tank. Again, sensors on the car will indicate if such condition would be the case. Therefore, we can say that an autonomous vehicle would be able to secure that it is in a good shape. Then we have the <u>situation</u>. During the training of the machine learning algorithms as described above the autonomous vehicle have been exposed to a number of different situations and thus also learned how to adapt the driving in these situations in order to drive safely. Therefore, we can say that an autonomous vehicle can handle different traffic, road and weather conditions etc. in a good and safe manner. Therefore, it seems we can conclude that an autonomous vehicle have competence! If it were to exercise that competence in an attempt to drive safely from work to home, it would with good reliability succeed in such attempt that would then be both <u>accurate</u> and <u>adroit</u> and thus <u>apt</u>!

So far so good! However, now we have to address the challenge of first and second order of knowledge. Our human driver was able to reflect and pass judgement based on his or her competence as to whether it would be safer to drive or to walk under the given circumstances. Would our autonomous vehicle be able to do so as well? What would be needed for an AI agent in order to be able to reflect, as a human would do? Well, an autonomous vehicle would be able based on its sensors to conclude that its shape would be unfit for safe driving, but would this be to reflect or just perceive though its cognitive faculties (sensors)? In addition, it would not be able to reflect on its skill, as it would not have anything to compare with. As for the situation, it would adapt its driving, but would it reflect on if the situation would be unfit for safe driving? If a passenger would tell it to drive home it would attempt to do so! Therefore, even if an autonomous vehicle would probably be more competent in driving compared to a human driver, it would not be able to reflect like a human would, and thus it would not be able to achieve full aptness on the second order. Why is this? Well it would have the experience from its training and it could be said to some extent to have an understanding but would not be able to pass judgement in an apt way.

Is there any way around this seeming dead end for machines ability to know? Well, one could of course discard the virtue epistemology definition of knowledge and instead take a pure empiricist view on knowledge in which case knowledge would only be the product of experience, which we have said that machines have, and sense perception that machines also can be said to have! Another approach would be to say as Linda Zagzebski does that Gettier problems are inescapable (Anthology, Sosa et al, 2008, chapter 17, pp.207-212) in which case we end up in a skeptic position where justified true beliefs do not hold and we cannot be sure, or possess knowledge, of anything. This of course will not help us since in this scenario machines will not know anything either. Yet another way would be to argue that what virtue epistemology defines as knowledge on the second reflective order is not really knowledge at all, but instead wisdom, in which case we only have knowledge on the first order where we have concluded that machines also know and possess knowledge!

## Will machines be able to know in the future?

Now, we have only examined one example above, but that example seem to indicate that today, at least from a virtue epistemology perspective, we have a challenge when it comes to machines abilities to reflect, and thus given the virtue epistemology definition of knowledge, also to know. So, what is the missing piece of the puzzle here? Sometimes when talking about reflecting we also add that we consciously reflect. Humans reflect when they are conscious, i.e. not unconscious like when sleeping. Is consciousness required in order to be able to reflect? If so, can a machine be conscious? Well, today the answer to that question is no, unless you support panpsychism, in which case the answer would be yes, but to a degree!

Will machines be able to achieve a conscious state[1]? Today we talk about that we are close to develop superintelligent AI. Bostrom defines superintelligence as "any intellect that greatly exceeds the cognitive performance of humans in virtually all domains of interest" (Bostrom, 2017, pp. 26). To that, he adds in the context of AI that it should be an intelligence in a machine substrate. Would such AI agent also be conscious? As soon as you start to talk about consciousness, every computer scientist and neuro scientist start to twitch! According to Stuart Russell, one of the leading persons in the field of artificial intelligence, consciousness does not matter! "All those Hollywood plots about machines mysteriously becoming conscious and hating humans are really missing the point: it's competence, not consciousness, that matters." (Russell, 2019, pp. 17). Well, it might not matter in terms of what an AI agent is capable of doing, however, it does matter for our inquiry about machines ability to know and to possess knowledge, since it seems to be a necessary component in order to be able to reflect. On the other hand, if knowledge is action and a superintelligent AI agent can greatly exceed the cognitive performance of humans in virtually all domains of interest, then maybe it is as Russell proposes that competence is more important than consciousness.

---

[1] For an interesting discussion on this topic please refer to (Fridman and Chalmers, 2020).

Alan Turing also brought up consciousness as one of the objections or critiques to his hypothesis that machines can think (Turing, 1950, pp. 11-12).

"This argument is very, well expressed in Professor Jefferson's Lister Oration for 1949, from which I quote. 'Not until a machine can write a sonnet or compose a concerto because of thoughts and emotions felt, and not by the chance fall of symbols, could we agree that machine equals brain-that is, not only write it but know that it had written it. No mechanism could feel (and not merely artificially signal, an easy contrivance) pleasure at its successes, grief when its valves fuse, be warmed by flattery, be made miserable by its mistakes, be charmed by sex, be angry or depressed when it cannot get what it wants.'" (Turing, 1950, pp. 11)

It should be noted though that even in the field of creative expression AI is actually doing quite some progress as can be seen in Marcus du Sautoy's book "The creativity code. How AI is learning to write, paint and think" (du Sautoy, 2019).

Now, Turing's defense against such critique is that in order to know if a machine could think one would have to be that machine, but this also goes for a human. In order to know that a man is conscious and thinking you would have to be that man. Turing then concludes that this is the solipsist point of view, which no one really want to accept (?). It comes close to Descartes reasoning that the only thing one can be certain about is the foundation *I* think, therefore *I* am (Descartes, 1641).

## Implications of transhumanism?

Now, there is another interesting path to explore when talking about "machine" intelligence and knowledge. That is the implications from transhumanism. A transhumanist sees the next step in human evolution to be technical augmentations to our human organism such that we will get different types of technical "superpowers" and more and more become like a machine (or Cyborg) (see more in Kurzweil, 2005).

If this happens the future superintelligent "machine" would still have a human component left to it, thus also still be conscious and able to consciously reflect, and therefore be able to possess fully apt knowledge! Maybe this path is one that would allow us to affirm that machine would be able to know.

## Implications for human knowledge?

So what have our inquiry into whether or not machines can know given us in terms of understanding our human ability to know? It seem that we as humans are gifted with consciousness and that this is the element that is needed in order for us to be able to know and possess knowledge in a fully apt reflective way based on judgement on the second order. However, one could ask if this conscious reflective ability is really knowledge. Passing judgement on the second order is based on our experience and understanding. This is something that we have accumulated over time so that we can reflect on our justified (true) beliefs and decide based on our first order knowledge what actions to take in order to competently succeed and be accurate in our attempt to alethically affirm something – to be fully apt in doing so. Is this not wisdom rather than knowledge? According to Stanford Encyclopedia of Philosophy (SEP, 2020), wisdom can be view from a number of perspectives, one being wisdom as knowledge where knowledge is said to be a necessary condition for wisdom, but it is then not the same as wisdom rather a constituent part of wisdom. This would imply that human knowledge would only reside on the first order (the animal level), and on the second order we have wisdom rather than knowledge.

## Conclusions

It would then seem that machines, or AI agents, at least today, cannot according to the virtue epistemology definition of knowledge be said to be able to know since an AI agent lack the ability to consciously reflect and thus cannot be said to be fully apt in its epistemic actions and alethic affirmations. Granted, we would be able to say that a machine would possess knowledge of the first order, but not as noted know full well by full aptness.

The missing component then for achieving knowledge of the second order seem to be consciousness, something that we as humans assume that we have, but which we again as humans assume a machine not to have, at least not in the same fashion as humans which would be needed for conscious reflection. At the same time, this notion could be challenged and we could argue that second order "knowledge" is not really knowledge but wisdom instead, in which case we would be able to conclude that also machines are able to know and possess knowledge!

The question though becomes what value we should assign to knowledge from a human perspective if in the future a superintelligent AI agent will be able to greatly exceed the cognitive performance of humans in virtually all domains of interest. What will our human ability to know be worth then? Maybe this is something that we again need to wait until the end of our current century to get an answer on.

**existentialrisk.ai**
Raymond Anneborg